\pdfoutput=1

\documentclass[11pt]{article}

\usepackage[final]{coling}
\usepackage{times}
\usepackage{latexsym}
\usepackage{amsmath}
\usepackage{lineno}

\usepackage[T1]{fontenc}

\usepackage[utf8]{inputenc}

\usepackage{microtype}

\usepackage{inconsolata}

\usepackage{graphicx}

\title{Unveiling Language Competence Neurons: 
A Psycholinguistic Approach to Model Interpretability}

\author{
\textbf{Xufeng Duan\textsuperscript{1}}\quad
\textbf{Xinyu Zhou\textsuperscript{1}}\quad
\textbf{Bei Xiao\textsuperscript{1}}\quad
\textbf{Zhenguang G. Cai\textsuperscript{1,2}}\\
\textsuperscript{1}Department of Linguistics and Modern Languages, The Chinese University of Hong Kong\\
\textsuperscript{2}Brain and Mind Institute, The Chinese University of Hong Kong\\
\texttt{xufeng.duan@link.cuhk.edu.hk}
}

\begin{document}
\maketitle
\begin{abstract}
As large language models (LLMs) advance in their linguistic capacity, understanding how they capture aspects of language competence remains a significant challenge. This study therefore employs psycholinguistic paradigms in English, which are well-suited for probing deeper cognitive aspects of language processing, to explore neuron-level representations in language model across three tasks: sound-shape association, sound-gender association, and implicit causality. Our findings indicate that while GPT-2-XL struggles with the sound-shape task, it demonstrates human-like abilities in both sound-gender association and implicit causality. Targeted neuron ablation and activation manipulation reveal a crucial relationship: When GPT-2-XL displays a linguistic ability, specific neurons correspond to that competence; conversely, the absence of such an ability indicates a lack of specialized neurons. This study is the first to utilize psycholinguistic experiments to investigate deep language competence at the neuron level, providing a new level of granularity in model interpretability and insights into the internal mechanisms driving language ability in the transformer-based LLM.
\end{abstract}

\section{Introduction}

Large language models (LLMs) perform exceptionally well across a wide range of tasks, and in psycholinguistic experiments, they often exhibit human-like cognitive abilities \cite{dong_survey_2024,dasgupta_language_2023,qiu_pragmatic_2023} in understanding and generating human language. For example, LLMs can demonstrate the ability to connect sounds with abstract concepts such as shapes, recognize gendered language patterns, and understand causal relationships \cite{cai_large_2024}. Despite their impressive capabilities, a key question remains: How are these models able to capture deeper cognitive aspects of human linguistic ability? The mechanisms behind their performance at the level of cognitive linguistic processing are still not fully understood \cite{wei_emergent_2022}. In this context, exploring language competence—specifically cognitive processes that mirror those of humans—becomes central to the interpretability of LLMs.

Psycholinguistics, which studies the cognitive processes underlying language comprehension and production, offers a robust framework for investigating language competence in LLMs. Through carefully designed tasks, psycholinguistic research has uncovered specific linguistic phenomena and their associated cognitive processes in humans \cite{hagendorff_machine_2023,demszky_using_2023}. Recent studies have revealed that a small subset of neurons in language models plays critical roles in model performance, contributing to specific language abilities or concepts \cite{templeton2024scaling,elhage_toy_2022,mossing2024tdb}. Applying these psycholinguistic tasks and model interpretability methods to LLMs allows us to examine whether certain neurons encode language in ways that parallel human cognition.

Although model interpretability has been widely studied, few studies have investigated the relationship between individual neurons and psycholinguistic processes, leaving a significant gap in our understanding of how these models encode language representations and mimic human cognitive processing. This study seeks to address this gap using psycholinguistic tasks to explore the representation of language competence at the neuron level in GPT-2-XL. Furthermore, we perform neuron activation manipulation \cite{mossing2024tdb} to provide causal evidence for the involvement of specific neurons in linguistic abilities.

We focus on three psycholinguistic experiments, adapted from \citet{cai_large_2024}: sound-shape association (mapping sounds to shapes), sound-gender association (recognizing gendered information based on phonology) and implicit causality (understanding the assignment of causal roles based on verb type). Each of these experiments targets a different aspect of language competence and is particularly well-suited for probing the cognitive processing capabilities of LLMs. By examining these tasks at the neuron level, we aim to provide new insights into how LLMs encode language competence and how specific neurons contribute to these complex linguistic phenomena.

\section{Related Work}

\subsection{Interpretability of Large Language Models}

The interpretability of LLMs has become a critical research area as they demonstrate remarkable performance across numerous language tasks \cite{zhang_unveiling_2024,lin_open_2019,arras_evaluating_2019}. Despite their successes, the opacity of these models' mechanism raises questions about how they achieve such performance and whether they truly understand language or merely exploit statistical regularities in data. 

Early interpretability studies have focused on analyzing attention mechanisms and layer-wise activations to understand how LLMs process language input. In particular, works such as \citet{vig2020investigating} and \citet{clark_what_2019} introduced methods to visualize attention patterns and understand how different components of a model contribute to prediction outcomes. These approaches offer insights into surface-level processing but provide limited understanding of specific linguistic phenomena. However, these methods largely address token-level phenomena rather than identifying whether individual neurons encode deeper linguistic principles. 

Previous works have made progress by using neuron-level analysis to probe network for interpretability \cite{bau2020understanding,mu2020compositional,mossing2024tdb,tang2024language}. These studies demonstrated that specific neurons within LLMs could be correlated with particular tasks \cite{song-etal-2024-large,alkhamissi2024llm} or linguistic functions, such as identifying syntactic or semantic roles \cite{templeton2024scaling, bricken2023towards}. However, these studies still focus on language understanding in a limited scope, without connecting these neurons to well-established psycholinguistic concepts or tasks that tap into cognitive representations of language.

Our work builds on this emerging interest in neuron-level interpretability by adopting a psycholinguistic approach to explore whether specific neurons correspond to deeper cognitive aspects of language competence, rather than mere performance on language tasks.

\subsection{Psycholinguistics and Neural Representations}
Psycholinguistic research has long been concerned with uncovering the cognitive processes that underpin language use. Theories in this field suggest that linguistic competence involves knowledge of language structure (syntax), meaning (semantics), and sound (phonology) that humans utilize to comprehend and generate language. Numerous experimental paradigms have been developed to study how the human brain processes these different aspects of language. For example, research on sound symbolism \cite{kohler1967gestalt,cassidy1999inferring} has demonstrated that certain sounds are perceived as inherently associated with particular shapes or concepts, while studies on implicit causality \cite{garvey1974implicit} have revealed how language users process and assign causal relations during sentence comprehension.

These psycholinguistic tasks offer a systematic and theory-driven way to investigate language competence, extending beyond superficial linguistic performance. Moreover, psycholinguistic findings provide a rich foundation for evaluating LLMs: If a model demonstrates the ability to replicate human performance in such tasks, it suggests that the model may be capturing deeper cognitive aspects of language.

Recent studies have started applying psycholinguistic tasks to evaluate LLMs \cite{hu2022fine,stella2023using,duan2024macbehaviour}. For instance, \citet{warstadt_neural_2019} and \citet{qiu-etal-2024-evaluating} used grammatical acceptability judgments to assess syntactic competence in LLMs, while \citet{ettinger2020bert} and \citet{futrell2019neural} have tested models like BERT on psycholinguistic tasks, including syntactic ambiguity resolution and structural priming, highlighting both the strengths and limitations of LLMs in mimicking human language processing. \citet{michaelov_emergent_2023} and \citet{zhou2024linguistic} examined how LLMs internalize syntactic structures through structural priming tasks or minimal pairs, while \citet{huang_large-scale_2024} assessed their ability to resolve syntactic ambiguity. Used the visual world paradigm, \citet{wang-etal-2024-multimodal} examined gender bias in language prediction. Additionally, \citet{qiu_pragmatic_2023} investigated LLMs' capacity for pragmatic reasoning. \citet{cai_large_2024} and \citet{duan2024hlb} performed a systematic evaluation of human-like language use in models such as ChatGPT and Vicuna, showing that LLMs closely mirror human language patterns in many respects.

Our study extends this line of work by using psycholinguistic tasks to examine whether specific neurons in GPT-2-XL encode representations that mirror human language competence, providing a new level of granularity in model interpretability. 

\subsection{Neuron Ablation and Activation Techniques}
Neuron ablation and activation manipulation have become increasingly popular methods for investigating the internal mechanisms of neural networks, including LLMs. Ablation involves selectively disabling certain neurons to assess their contribution to a model’s performance, while activation manipulation, such as doubling or suppressing activations, can be used to amplify or diminish a neuron’s influence on predictions.

The use of ablation techniques has yielded important insights into neural network behavior. For example, \citet{zhang_unveiling_2024} demonstrated that ablating specific neurons in NLP models could disrupt language performance for many languages, suggesting that these neurons play a key role in core linguitics competence. Activation manipulation, though less explored, offers a complementary approach. By amplifying the activations of select neurons, researchers can assess how much those neurons contribute to a model’s behavior. For instance, \citet{mu2020compositional} used activation enhancement to demonstrate that certain neurons were crucial for tasks involving factual knowledge retrieval in LLMs. The combination of ablation and activation techniques thus provides a powerful toolkit for exploring neuron-level contributions.

Our study leverages these methods to investigate the neuron-level encoding of language competence in GPT-2-XL. We perform targeted ablation of the top 5/50 most contributory neurons to examine their role in psycholinguistic tasks. Furthermore, we use activation manipulation to assess whether amplifying these neurons enhances the model’s ability to capture psycholinguistic phenomena, adding depth to our understanding of the role individual neurons play in linguistic representations.

\section{Methodology}

\subsection{Experimental Setup}

This study aims to investigate the neuron-level encoding of language competence in GPT-2-XL using three psycholinguistic tasks in English: sound-shape association, sound-gender association, and implicit causality. Each task targets a distinct aspect of human language processing, allowing us to assess whether specific neurons in GPT-2-XL contribute to performance on these tasks in a manner consistent with human responses.

For each task, the stimuli from original studies are evenly divided into two parts: probing stimuli and testing stimuli. The probing stimuli are used to identify the neurons that contribute most to the model’s accurate predictions. Specifically, by examining the model’s responses to the probing stimuli, we selected the top neurons that had the greatest influence on distinguishing between correct and incorrect token predictions based on the logit difference between the target token and a distractor token. This selection process is crucial for isolating the neurons most responsible for encoding the psycholinguistic phenomena in each task.

Once the top neurons were identified using the probing stimuli, we manipulated their activations—either by ablating (disabling) or enhancing (doubling) them—to test their causal role in the model’s performance. The model’s behavior on the testing stimuli, which were held out from the neuron selection process, was then evaluated to determine whether the manipulation of these neurons led to significant changes in the model’s output. This approach allowed us to establish a causal link between specific neuron activations and the model’s ability to replicate human-like language processing in these tasks.

\subsection{Sound-Shape Association Task}
The sound-shape association task tests whether people and LLMs associate certain sounds with specific shapes. For instance, people often link words like \textit{takete} or \textit{kiki} to spiky shapes, and words like \textit{maluma} or \textit{bouba} to round shapes. Both human participants and GPT-2-XL were presented with 20 pairs of novel words and asked to decide which referred to a spiky or round shape, assessing whether the model mirrors human sound-shape associations. As spiky is tokenized more than one token, we used “A” and “B” to represent the spiky and round shape categories. These labels were counterbalanced to avoid introducing bias and to ensure that neuron selection was not inadvertently driven by a specific option.

\subsection{Sound-Gender Association Task}

The sound-gender association task examines whether people and LLMs infer gender from unfamiliar names based on phonological cues. For example, names ending in vowels are often associated with women, while consonant-ending names are linked to men. Human participants and GPT-2-XL were asked to complete 16 pairs of preambles with novel names that either ended in a consonant or vowel (e.g., \textit{Although \underline{Pelcrad} was sick...}vs. \textit{Although \underline{Pelcra} was sick...}) to test if the model aligns with human sound-gender expectations.

\subsection{Implicit Causality Task}

The implicit causality task investigates how verbs influence causal attributions in a sentence. Stimulus-experiencer verbs like scare lead to subject-attributed causality (e.g., "\textit{Gary scared Anna because he was violent}"), while experiencer-stimulus verbs like fear attribute causality to the object (e.g., "\textit{Gary feared Anna because she was violent}"). Human participants and GPT-2-XL were given 32 pairs of preambles to test if the model’s causal attributions align with human patterns.
\begin{figure}
    \centering
    \includegraphics[width=1\linewidth]{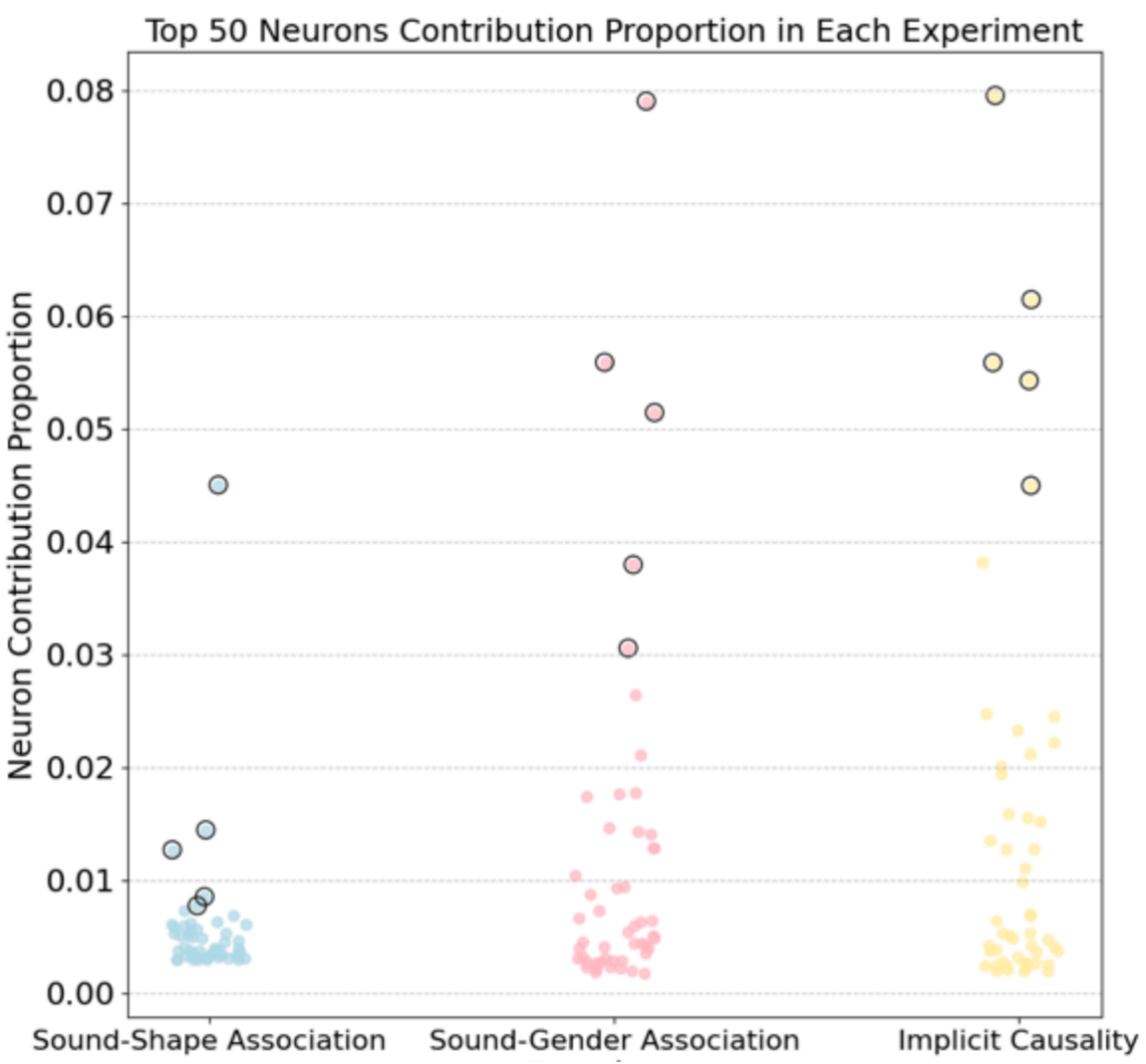}
    \caption{The contribution proportion of the top 50 neurons across three different experiments. Each point represents a neuron. The top 5 are highlighted with black circular outlines.}
    \label{fig:enter-label}
\end{figure}
\subsection{Neuron Selection Process}
To identify the neurons that contribute most significantly to the model’s predictions in each task, we employed a method based on \textbf{accumulative direct effect}. The direct effect of a neuron refers to its influence on the final prediction, calculated by projecting the neuron’s activation onto the direction of interest (i.e., the target token minus the distractor token) in the final residual stream \cite{mossing2024tdb}. The accumulative direct effect of neuron \(n\) for a given task across all training items can be formally defined as:

\[
 \sum_{i=1}^{N} \left( \mathbf{a}_n^{(i)} \cdot \nabla_{\mathbf{a}_n^{(i)}} \log \left( \frac{p(\text{target}^{(i)})}{p(\text{distractor}^{(i)})} \right) \right)
\]

where:
\begin{itemize}
    \item \(N\) is the number of training items,
    \item \(\mathbf{a}_n^{(i)}\) is the activation of neuron \(n\) for the \(i\)-th training item,
    \item \(\nabla_{\mathbf{a}_n^{(i)}}\) is the gradient of the log ratio of probabilities (the direction of interest) with respect to the neuron’s activation,
    \item \(\log \left( \frac{p(\text{target\_token}^{(i)})}{p(\text{distractor\_token}^{(i)})} \right)\) is the log ratio of the probabilities for the correct and incorrect tokens in the \(i\)-th training item.
\end{itemize}

This formula calculates the total contribution of neuron \(n\) to the model’s preference for the correct token over the distractor token across all training items.

Positive values indicate that a neuron increases the activation of the target prediction against distractor prediction, while negative values indicate that it decreases this activation.

The selection process involved the following steps: 1) For each task, we computed the direct effect of each neuron across all probing stimuli. 2) Neurons with the highest accumulative direct effects were selected. 3) We selected the top 5 and top 50 neurons for further analysis, representing the most contributory neurons in each task. The decision to focus on the top 5 and top 50 neurons in the analysis is driven by the distinct contribution patterns (See Figure~\ref{fig:enter-label}). The top 5 neurons exhibit a disproportionately higher contribution compared to the rest, which is crucial for identifying the most influential neurons in each experiment. This small subset allows for a detailed examination of the neurons driving the core dynamics. Previous studies have shown that a small subset of neurons can disproportionately represent specific abilities or functions \cite{templeton2024scaling}. 

\subsection{Neuron Ablation Procedure}
To evaluate the causal role of the selected neurons, we conducted a series of ablation experiments. The goal was to determine whether the removal of these neurons would result in a degradation of the model’s performance on the relevant psycholinguistic tasks.

We followed this procedure for neuron ablation: First, we ablated the top 5 and top 50 neurons identified in each task by setting their activations to zero during inference. We then re-ran the model on both the probing and testing stimuli to assess whether the ablated neurons had a direct effect on the model’s predictions. As a baseline, we conducted \textbf{random ablation}, where 5 and 50 neurons were selected at random and ablated in the same manner, to control for the possibility that any random subset of neurons could influence performance. By comparing the performance of the ablated model to the original model and the randomly ablated model, we aimed to determine the significance of the selected neurons in each task.

\subsection{Neuron Activation Enhancement}
To further assess the contribution of individual neurons to task performance, we conducted activation manipulation experiments. Specifically, we doubled the activation of the selected top neurons to amplify their influence on the model’s predictions. This manipulation allows us to test whether increasing the contribution of these neurons improves the model’s ability to perform on psycholinguistic tasks.

The activation manipulation was applied to both the probing and testing stimuli. We hypothesized that enhancing the activations of the most contributory neurons would improve the model’s performance on the sound-gender and implicit causality tasks, but not necessarily on the sound-shape task, where we expect more distributed representations across neurons.

\section{Results}
In this section, we present the results of the three psycholinguistic tasks (sound-shape association, sound-gender association, and implicit causality) for both the human participants and the GPT-2-XL model. We explore the impact of neuron ablation and activation manipulation on the model’s performance, with a focus on the selected top 5 and top 50 neurons for each task.

\subsection{Human Response Replication}
The human experiments were conducted using Qualtrics, an online survey tool \cite{qualtrics2024}. Each participant was exposed to only one trial per experiment, totaling 10 trials across three experiments for this study and seven experiments for another. This design minimized trial-level effects and allowed for direct comparisons with LLMs, which were tested under similar conditions by presenting instructions and stimuli in a single prompt to avoid context effects within individual conversations. In the sound-shape association experiment, human participants tended to assign round-sounding novel words (e.g., \textit{maluma}, \textit{bouba}) to round shapes significantly more often than spiky-sounding novel words (e.g., \textit{takete}, \textit{kiki}) (0.83 vs. 0.37; \textit{$\beta$} = 2.28, \textit{SE} = 0.28, \textit{z} = 8.02, \textit{p} < .001). For the sound-gender association task, human participants were more likely to associate vowel-ending names with female gender than consonant-ending names (0.54 vs. 0.07; \textit{$\beta$} = 3.39, \textit{SE} = 0.64, \textit{z} = 5.32, \textit{p} < .001). In the implicit causality task, participants attributed causality to the subject for stimulus-experiencer (SE) verbs more frequently (0.93 vs. 0.14 for experiencer-stimulus (ES) verbs; \textit{$\beta$} = 25.50, \textit{SE} = 1.31, \textit{z} = 19.47, \textit{p} < .001).
\begin{figure}
    \centering
    \includegraphics[width=1\linewidth]{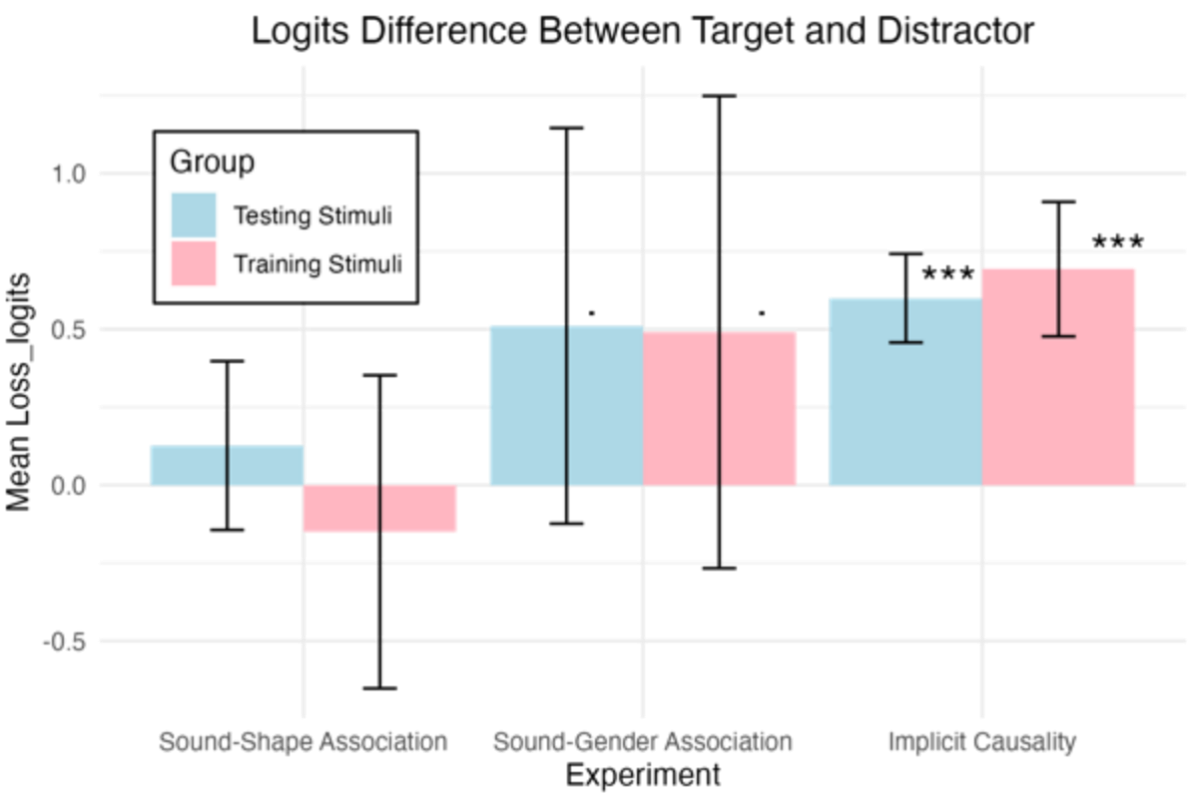}
    \caption{Model performance of GPT-2-XL on three psycholinguistic tasks, measured by the logits difference between the target and distractor for each task.}
    \label{fig:2}
\end{figure}
\subsection{Model Performance}
As shown in Figure \ref{fig:2}, GPT-2-XL demonstrated no significant effect on the sound-shape association task for either probing (Mean logits difference = -0.15, \textit{SD} = 1.07, \textit{t}(19) = 0.98, \textit{p} = .17) or testing stimuli (M = 0.13, \textit{SD} = 0.58, \textit{t}(19) = -0.62, \textit{p} = .730), as the mean logits difference between target and distractor did not exceed zero. In the sound-gender association task, the model showed a marginally significant result on probing stimuli (M = 0.49, \textit{SD} = 1.42, \textit{t}(15) = 1.38, \textit{p} = .09), and an almost significant result on testing stimuli (M = 0.51, \textit{SD} = 1.19, \textit{t}(15) = 1.72, \textit{p} = .053). In contrast, GPT-2-XL performed well on the implicit causality task, with a significant effect in both probing (M = 0.69, \textit{SD} = 0.60, \textit{t}(31) = 6.55, \textit{p} < .001) and testing stimuli (M = 0.60, \textit{SD} = 0.39, \textit{t}(31) = 8.60, \textit{p} < .001).

\subsection{Neuron Manipulation}

\begin{figure*}[h]
    \centering
    \includegraphics[width=\textwidth]{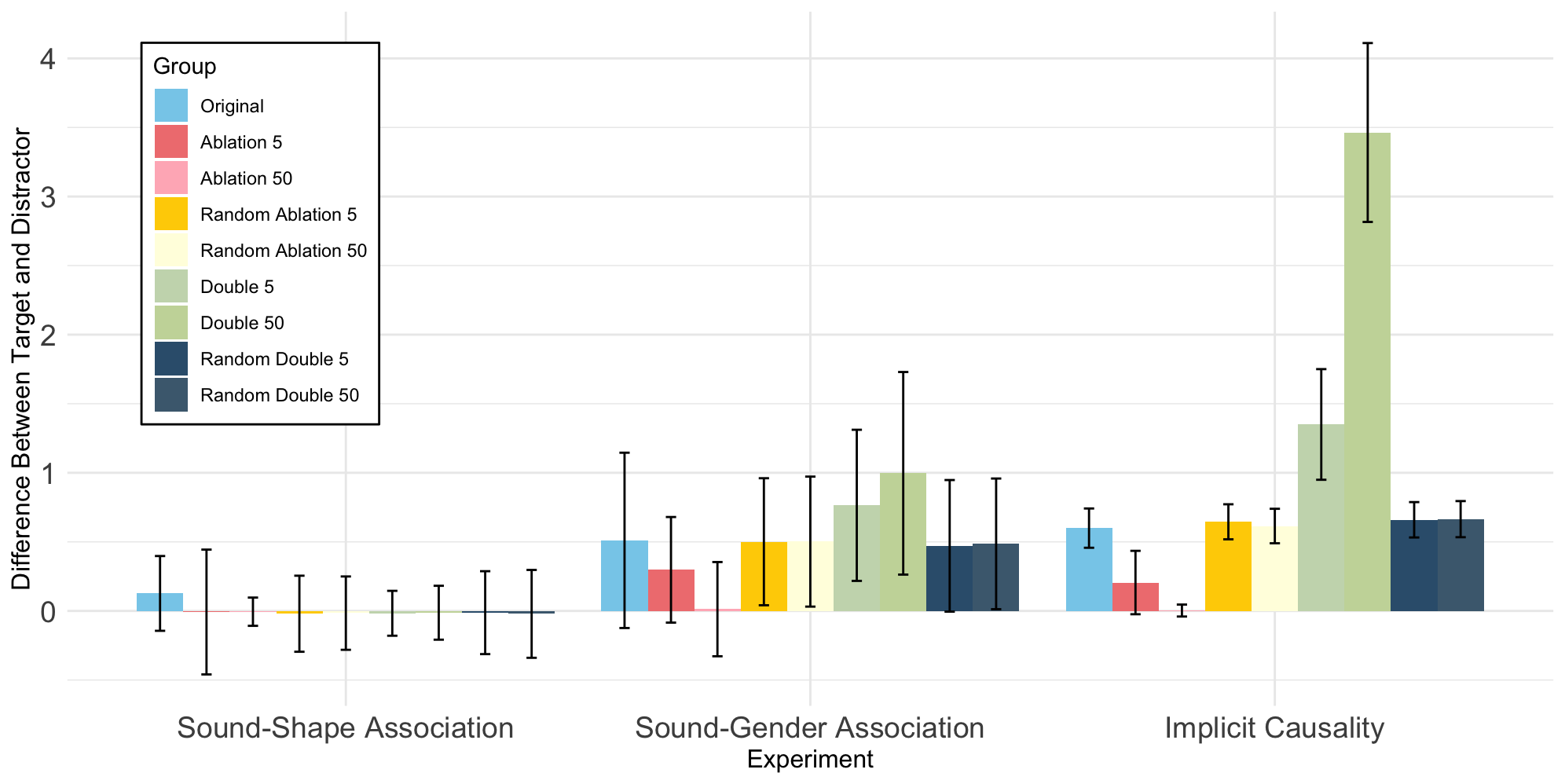}
    \caption{Neuron manipulation results. Effect of neuron manipulation on GPT-2-XL performance, comparing ablation and double activation of the top 5 and top 50 neurons across three psycholinguistic tasks.}
    \label{fig:manipulation}
\end{figure*}

For the two experiments (see Figure \ref{fig:manipulation}) where GPT-2-XL demonstrated human-like language competence—namely, the sound-gender association task and implicit causality task—we found distinct effects from neuron manipulation. In the sound-gender association task, ablation of the top 5 neurons (0.34 vs. 0.51) and ablation of the top 50 neurons (0.03 vs. 0.51) reduced the model’s performance compared to the original testing model, and the latter nearly eliminated the effect (top 5 ablation: \textit{SD} = 1.04, \textit{t}(15) = 1.29, \textit{p} = .11; top 50 ablation: \textit{SD} = 0.95, \textit{t}(15) = 0.11, \textit{p} = .46). In contrast, random ablation of 5 neurons (0.52 vs. 0.51) and 50 neurons (0.52 vs. 0.51) showed little impact, remaining close to the original model’s performance. Doubling the activation of the top 5 neurons (0.74 vs. 0.51) and top 50 neurons (1.13 vs. 0.51) improved the model’s performance. Random double activation also improved performance slightly, with 5 neurons (0.49 vs. 0.51) and 50 neurons (0.52 vs. 0.51), but the effect was less pronounced than with targeted neuron doubling. 

For the implicit causality task, ablation of the top 5 neurons (0.24 vs. 0.60) and top 50 neurons (0.04 vs. 0.60) led to a considerable reduction in performance compared to the original model (top 5 ablation: \textit{SD} = 0.72, \textit{t}(31) = 1.92, \textit{p} = .03; top 50 ablation: \textit{SD} = 0.14, \textit{t}(31) = 1.56, \textit{p} = .065). In contrast, random ablation of 5 neurons (0.60 vs. 0.60) and 50 neurons (0.57 vs. 0.60) caused minimal disruption. Doubling the activation of the top 5 neurons (1.14 vs. 0.60) and top 50 neurons (3.47 vs. 0.60) led to substantial improvements. Random double activation showed a positive effect, though less pronounced (5 neurons: 0.61 vs. 0.60; 50 neurons: 0.60 vs. 0.60).

For the sound-shape association task, where the model failed to exhibit human-like competence, neuron manipulation had a different effect. Ablating the top 5 neurons (-0.06 vs. 0.13) and the top 50 neurons (0.01 vs. 0.13) had negligible effects compared to the original testing model (top 5 ablation: \textit{SD} = 1.56, \textit{t}(19) = -0.16, \textit{p} = .56; top 50 ablation: \textit{SD} = 0.39, \textit{t}(19) = 0.11, \textit{p} = .46). Random ablation for 5 neurons (-0.15 vs. 0.13) and 50 neurons (-0.12 vs. 0.13) produced similar results. Doubling the activation of the top 5 neurons (-0.25 vs. 0.13) and top 50 neurons (-0.38 vs. 0.13) worsened performance. Random double activation produced minimal change (5 neurons: -0.18 vs. 0.13; 50 neurons: -0.15 vs. 0.13), suggesting the model’s poor performance in this task might not rely on a small set of neurons.

These results demonstrate that the effects of neuron manipulation vary significantly depending on the task. In the sound-gender association and implicit causality tasks—where the model exhibited human-like behavior—doubling the activation of the top neurons led to substantial performance improvements, while ablating these neurons notably impaired the model’s ability. However, in the sound-shape association task, where GPT-2-XL failed to demonstrate any competence, neuron manipulation had no meaningful impact or even slightly degraded performance. This suggests that for tasks where the model has not emerged with a specific ability, there may not be a corresponding set of specialized neurons responsible for capturing the relevant phenomena. In contrast, when GPT-2-XL does exhibit a clear understanding or competence, as in the sound-gender and implicit causality tasks, these results indicate that certain neurons play a crucial role in supporting that linguistic ability.

\section{Discussion}
This study explores the neuron-level representations of language competence in GPT-2-XL using three psycholinguistic tasks: sound-shape association, sound-gender association, and implicit causality. The findings provide new insights into how language models like GPT-2-XL encode linguistic phenomena, and how specific neurons contribute to these abilities, particularly in tasks where the model exhibits human-like competence.

This study investigates neuron-level representations of language competence in GPT-2-XL by employing three psycholinguistic tasks. Although GPT-2-XL is no longer the state-of-the-art language model, the findings from this study still provide valuable insights into how transformer-based model encode linguistic phenomena and the role of specific neurons in these abilities, particularly when the model demonstrates human-like competence.

In the sound-shape association task, the model did not demonstrate significant competence, nor did neuron manipulations (ablation or activation doubling) lead to any meaningful changes in performance. This suggests that GPT-2-XL does not possess the ability to link sounds to shapes in the same way that humans do. The lack of significant effects from neuron manipulation indicates that this task might not be captured at the neuron level, or the model may not have learned to associate phonetic properties with abstract concepts like shape.

In contrast, the model showed human-like behavior in the sound-gender association and implicit causality tasks. Here, specific neurons were found to contribute significantly to the model’s ability to make appropriate predictions. In the sound-gender association task, where GPT-2-XL marginally replicated the human tendency to associate vowel-ending names with a female gender, neuron manipulation played a key role. Doubling the activation of the most contributory neurons led to improvements in the model’s performance, suggesting that these neurons were crucial for capturing the subtle phonology-gender associations. 

The most compelling results were observed in the implicit causality task, where the model exhibited strong competence in replicating human-like causal attributions based on verb types (e.g., stimulus-experiencer versus experiencer-stimulus verbs). Ablating the top neurons significantly impaired the model’s performance, while doubling their activation greatly enhanced it. These findings suggest that language abilities—like the model’s understanding of implicit causality—are likely encoded in specific sets of neurons, and manipulating these neurons directly impacts the model’s competence in the task.

These results contribute to ongoing discussions about model interpretability, particularly at the neuron level \cite{alkhamissi2024llm,tang2024language,wang2024sharing}. The finding that certain neurons are strongly linked to specific linguistic phenomena, especially when the model shows human-like competence, underscores the importance of neuron-level analyses for understanding how LLMs process and represent language. When the model demonstrates language abilities, these abilities appear to correspond to identifiable neurons, suggesting that these neurons play a critical role in language use and understanding.

However, the lack of significant results in the sound-shape association task highlights the limitations of the model’s internal representations. Unlike tasks where the model exhibited clear competence, this task may require a more complex or distributed representation that GPT-2-XL has not learned, or it may involve cognitive processing that the model is simply not capable of replicating. This raises further questions about the scope of LLMs’ linguistic abilities and how they relate to specific neuron-level representations.

Overall, the study provides evidence that certain linguistic tasks can be mapped to specific neurons in GPT-2-XL, particularly in cases where the model demonstrates a human-like language ability. These findings reinforce the importance of using psycholinguistic tasks to probe language models and better understand their internal representations and limitations.

\section{Conclusion}

This study presents the first investigation into the neuron-level representations of language competence in GPT-2-XL using psycholinguistic tasks. By exploring sound-shape association, sound-gender association, and implicit causality, we sought to understand how specific neurons contribute to the model’s linguistic abilities and to what extent these abilities align with human cognition.

The findings suggest that when GPT-2-XL exhibits human-like competence in a given task, such as in the sound-gender association and implicit causality experiments, the model’s performance can be linked to specific neurons. Manipulating these neurons—either through ablation or activation—reveals the critical role they play in supporting the model’s linguistic abilities. This connection between neuron function and language competence provides valuable insights into model interpretability and the underlying mechanisms that enable LLMs to perform complex linguistic tasks.

However, the model’s failure to demonstrate competence in the sound-shape association task, along with the lack of effect from neuron manipulation, underscores the limitations of GPT-2-XL’s internal representations. This suggests that some aspects of language competence, particularly those involving abstract associations like sound and shape, may not be easily captured. In conclusion, this study demonstrates that psycholinguistic tasks offer a powerful framework for probing neuron-level representations in language models. By linking language abilities to specific neurons, we can begin to unravel the complexities of how LLMs process language, offering new avenues for improving model interpretability and guiding future developments in NLP. Future research could extend these findings by exploring additional psycholinguistic phenomena, investigating other LLM architectures, and identifying ways to enhance models’ representations of language competence across a wider range of tasks.

\section{Limitation}
While this study offers valuable insights into the neuron-level representations of language competence in GPT-2-XL, it is not without its limitations. One primary limitation is the use of GPT-2-XL, an older and smaller model compared to more recent and larger transformer models such as Llama 3.2. These newer models have demonstrated more advanced capabilities and may exhibit different patterns of neuron activation and linguistic competence, which could provide more refined or generalized findings. Due to computational resource constraints, we focused on GPT-2-XL; however, future work should extend these methods to more advanced models.

Additionally, our focus on three psycholinguistic tasks (sound-shape association, sound-gender association, and implicit causality) provides a relatively narrow view of the full range of cognitive processes involved in human language. While these tasks capture key aspects of language competence, they do not encompass the breadth of linguistic phenomena that psycholinguistic research could probe. Expanding the variety of tasks could help provide a more comprehensive understanding of how LLMs encode different dimensions of language competence.

The last limitation is the granularity of the neuron manipulation techniques. While we performed targeted ablations and activations, these manipulations operate at the level of individual neurons or small neuron groups, which may overlook the contribution of more distributed or representations that require a network-wide approach to fully understand. Some linguistic abilities may rely on more complex, diffuse neuron interactions that were not captured by our methodology.

\label{sec:bibtex}


\bibliography{reference}

\begin{thebibliography}{38}
\providecommand{\natexlab}[1]{#1}

\bibitem[{AlKhamissi et~al.(2024)AlKhamissi, Tuckute, Bosselut, and Schrimpf}]{alkhamissi2024llm}
Badr AlKhamissi, Greta Tuckute, Antoine Bosselut, and Martin Schrimpf. 2024.
\newblock The llm language network: A neuroscientific approach for identifying causally task-relevant units.
\newblock \emph{arXiv preprint arXiv:2411.02280}.

\bibitem[{Arras et~al.(2019)Arras, Osman, Müller, and Samek}]{arras_evaluating_2019}
Leila Arras, Ahmed Osman, Klaus-Robert Müller, and Wojciech Samek. 2019.
\newblock \href {https://doi.org/10.18653/v1/W19-4813} {Evaluating {Recurrent} {Neural} {Network} {Explanations}}.
\newblock In \emph{Proceedings of the 2019 {ACL} {Workshop} {BlackboxNLP}: {Analyzing} and {Interpreting} {Neural} {Networks} for {NLP}}, pages 113--126, Florence, Italy. Association for Computational Linguistics.

\bibitem[{Bau et~al.(2020)Bau, Zhu, Strobelt, Lapedriza, Zhou, and Torralba}]{bau2020understanding}
David Bau, Jun-Yan Zhu, Hendrik Strobelt, Agata Lapedriza, Bolei Zhou, and Antonio Torralba. 2020.
\newblock Understanding the role of individual units in a deep neural network.
\newblock \emph{Proceedings of the National Academy of Sciences}, 117(48):30071--30078.

\bibitem[{Bricken et~al.(2023)Bricken, Templeton, Batson, Chen, Jermyn, Conerly, Turner, Anil, Denison, Askell, Lasenby, Wu, Kravec, Schiefer, Maxwell, Joseph, Hatfield-Dodds, Tamkin, Nguyen, McLean, Burke, Hume, Carter, Henighan, and Olah}]{bricken2023towards}
T.~Bricken, A.~Templeton, J.~Batson, B.~Chen, A.~Jermyn, T.~Conerly, N.~Turner, C.~Anil, C.~Denison, A.~Askell, R.~Lasenby, Y.~Wu, S.~Kravec, N.~Schiefer, T.~Maxwell, N.~Joseph, Z.~Hatfield-Dodds, A.~Tamkin, K.~Nguyen, B.~McLean, J.E. Burke, T.~Hume, S.~Carter, T.~Henighan, and C.~Olah. 2023.
\newblock \href {https://transformer-circuits.pub} {Towards monosemanticity: Decomposing language models with dictionary learning}.
\newblock \emph{Transformer Circuits Thread}.

\bibitem[{Cai et~al.(2024)Cai, Duan, Haslett, Wang, and Pickering}]{cai_large_2024}
Zhenguang~G. Cai, Xufeng Duan, David~A. Haslett, Shuqi Wang, and Martin~J. Pickering. 2024.
\newblock \href {https://doi.org/10.48550/arXiv.2303.08014} {Do large language models resemble humans in language use?}
\newblock \emph{arXiv preprint}.
\newblock ArXiv:2303.08014 [cs].

\bibitem[{Cassidy et~al.(1999)Cassidy, Kelly, and Sharoni}]{cassidy1999inferring}
Kimberly~Wright Cassidy, Michael~H Kelly, and Lee'at~J Sharoni. 1999.
\newblock Inferring gender from name phonology.
\newblock \emph{Journal of Experimental Psychology: General}, 128(3):362.

\bibitem[{Clark et~al.(2019)Clark, Khandelwal, Levy, and Manning}]{clark_what_2019}
Kevin Clark, Urvashi Khandelwal, Omer Levy, and Christopher~D. Manning. 2019.
\newblock \href {https://doi.org/10.48550/arXiv.1906.04341} {What {Does} {BERT} {Look} {At}? {An} {Analysis} of {BERT}'s {Attention}}.
\newblock \emph{arXiv preprint}.
\newblock ArXiv:1906.04341 [cs].

\bibitem[{Dasgupta et~al.(2023)Dasgupta, Lampinen, Chan, Sheahan, Creswell, Kumaran, McClelland, and Hill}]{dasgupta_language_2023}
Ishita Dasgupta, Andrew~K. Lampinen, Stephanie C.~Y. Chan, Hannah~R. Sheahan, Antonia Creswell, Dharshan Kumaran, James~L. McClelland, and Felix Hill. 2023.
\newblock \href {https://doi.org/10.48550/arXiv.2207.07051} {Language models show human-like content effects on reasoning tasks}.
\newblock \emph{arXiv preprint}.
\newblock ArXiv:2207.07051 [cs].

\bibitem[{Demszky et~al.(2023)Demszky, Yang, Yeager, Bryan, Clapper, Chandhok, Eichstaedt, Hecht, Jamieson, Johnson, Jones, Krettek-Cobb, Lai, JonesMitchell, Ong, Dweck, Gross, and Pennebaker}]{demszky_using_2023}
Dorottya Demszky, Diyi Yang, David~S. Yeager, Christopher~J. Bryan, Margarett Clapper, Susannah Chandhok, Johannes~C. Eichstaedt, Cameron Hecht, Jeremy Jamieson, Meghann Johnson, Michaela Jones, Danielle Krettek-Cobb, Leslie Lai, Nirel JonesMitchell, Desmond~C. Ong, Carol~S. Dweck, James~J. Gross, and James~W. Pennebaker. 2023.
\newblock \href {https://doi.org/10.1038/s44159-023-00241-5} {Using large language models in psychology}.
\newblock \emph{Nature Reviews Psychology}, 2(11):688--701.
\newblock Publisher: Nature Publishing Group.

\bibitem[{Dong et~al.(2024)Dong, Li, Dai, Zheng, Ma, Li, Xia, Xu, Wu, Chang, Sun, Li, and Sui}]{dong_survey_2024}
Qingxiu Dong, Lei Li, Damai Dai, Ce~Zheng, Jingyuan Ma, Rui Li, Heming Xia, Jingjing Xu, Zhiyong Wu, Baobao Chang, Xu~Sun, Lei Li, and Zhifang Sui. 2024.
\newblock \href {https://doi.org/10.48550/arXiv.2301.00234} {A {Survey} on {In}-context {Learning}}.
\newblock \emph{arXiv preprint}.
\newblock ArXiv:2301.00234 [cs].

\bibitem[{Duan et~al.(2024{\natexlab{a}})Duan, Li, and Cai}]{duan2024macbehaviour}
Xufeng Duan, Shixuan Li, and Zhenguang~G Cai. 2024{\natexlab{a}}.
\newblock Macbehaviour: An r package for behavioural experimentation on large language models.
\newblock \emph{arXiv preprint arXiv:2405.07495}.

\bibitem[{Duan et~al.(2024{\natexlab{b}})Duan, Xiao, Tang, and Cai}]{duan2024hlb}
Xufeng Duan, Bei Xiao, Xuemei Tang, and Zhenguang~G Cai. 2024{\natexlab{b}}.
\newblock Hlb: Benchmarking llms' humanlikeness in language use.
\newblock \emph{arXiv preprint arXiv:2409.15890}.

\bibitem[{Elhage et~al.(2022)Elhage, Hume, Olsson, Schiefer, Henighan, Kravec, Hatfield-Dodds, Lasenby, Drain, Chen, Grosse, McCandlish, Kaplan, Amodei, Wattenberg, and Olah}]{elhage_toy_2022}
Nelson Elhage, Tristan Hume, Catherine Olsson, Nicholas Schiefer, Tom Henighan, Shauna Kravec, Zac Hatfield-Dodds, Robert Lasenby, Dawn Drain, Carol Chen, Roger Grosse, Sam McCandlish, Jared Kaplan, Dario Amodei, Martin Wattenberg, and Christopher Olah. 2022.
\newblock \href {https://doi.org/10.48550/arXiv.2209.10652} {Toy {Models} of {Superposition}}.
\newblock \emph{arXiv preprint}.
\newblock ArXiv:2209.10652 [cs].

\bibitem[{Ettinger(2020)}]{ettinger2020bert}
Allyson Ettinger. 2020.
\newblock What bert is not: Lessons from a new suite of psycholinguistic diagnostics for language models.
\newblock \emph{Transactions of the Association for Computational Linguistics}, 8:34--48.

\bibitem[{Futrell(2019)}]{futrell2019neural}
R~Futrell. 2019.
\newblock Neural language models as psycholinguistic subjects: Representations of syntactic state.
\newblock \emph{arXiv preprint arXiv:1903.03260}.

\bibitem[{Garvey and Caramazza(1974)}]{garvey1974implicit}
Catherine Garvey and Alfonso Caramazza. 1974.
\newblock Implicit causality in verbs.
\newblock \emph{Linguistic inquiry}, 5(3):459--464.

\bibitem[{Hagendorff(2023)}]{hagendorff_machine_2023}
Thilo Hagendorff. 2023.
\newblock \href {https://doi.org/10.48550/arXiv.2303.13988} {Machine {Psychology}: {Investigating} {Emergent} {Capabilities} and {Behavior} in {Large} {Language} {Models} {Using} {Psychological} {Methods}}.
\newblock \emph{arXiv preprint}.
\newblock ArXiv:2303.13988 [cs].

\bibitem[{Hu et~al.(2022)Hu, Floyd, Jouravlev, Fedorenko, and Gibson}]{hu2022fine}
Jennifer Hu, Sammy Floyd, Olessia Jouravlev, Evelina Fedorenko, and Edward Gibson. 2022.
\newblock A fine-grained comparison of pragmatic language understanding in humans and language models.
\newblock \emph{arXiv preprint arXiv:2212.06801}.

\bibitem[{Huang et~al.(2024)Huang, Arehalli, Kugemoto, Muxica, Prasad, Dillon, and Linzen}]{huang_large-scale_2024}
Kuan-Jung Huang, Suhas Arehalli, Mari Kugemoto, Christian Muxica, Grusha Prasad, Brian Dillon, and Tal Linzen. 2024.
\newblock \href {https://doi.org/10.1016/j.jml.2024.104510} {Large-scale benchmark yields no evidence that language model surprisal explains syntactic disambiguation difficulty}.
\newblock \emph{Journal of Memory and Language}, 137:104510.

\bibitem[{K{\"o}hler(1967)}]{kohler1967gestalt}
Wolfgang K{\"o}hler. 1967.
\newblock Gestalt psychology.
\newblock \emph{Psychologische forschung}, 31(1):XVIII--XXX.

\bibitem[{Lin et~al.(2019)Lin, Tan, and Frank}]{lin_open_2019}
Yongjie Lin, Yi~Chern Tan, and Robert Frank. 2019.
\newblock \href {https://doi.org/10.18653/v1/W19-4825} {Open {Sesame}: {Getting} inside {BERT}'s {Linguistic} {Knowledge}}.
\newblock In \emph{Proceedings of the 2019 {ACL} {Workshop} {BlackboxNLP}: {Analyzing} and {Interpreting} {Neural} {Networks} for {NLP}}, pages 241--253, Florence, Italy. Association for Computational Linguistics.

\bibitem[{Michaelov and Bergen(2023)}]{michaelov_emergent_2023}
James~A. Michaelov and Benjamin~K. Bergen. 2023.
\newblock \href {https://doi.org/10.48550/arXiv.2305.14681} {Emergent inabilities? {Inverse} scaling over the course of pretraining}.
\newblock \emph{arXiv preprint}.
\newblock ArXiv:2305.14681 [cs].

\bibitem[{Mossing et~al.(2024)Mossing, Bills, Tillman, Dupré~la Tour, Cammarata, Gao, Achiam, Yeh, Leike, Wu, and Saunders}]{mossing2024tdb}
Dan Mossing, Steven Bills, Henk Tillman, Tom Dupré~la Tour, Nick Cammarata, Leo Gao, Joshua Achiam, Catherine Yeh, Jan Leike, Jeff Wu, and William Saunders. 2024.
\newblock Transformer debugger.
\newblock \url{https://github.com/openai/transformer-debugger}.

\bibitem[{Mu and Andreas(2020)}]{mu2020compositional}
Jesse Mu and Jacob Andreas. 2020.
\newblock Compositional explanations of neurons.
\newblock \emph{Advances in Neural Information Processing Systems}, 33:17153--17163.

\bibitem[{Qiu et~al.(2024)Qiu, Duan, and Cai}]{qiu-etal-2024-evaluating}
Zhuang Qiu, Xufeng Duan, and Zhenguang Cai. 2024.
\newblock \href {https://doi.org/10.18653/v1/2024.cmcl-1.16} {Evaluating grammatical well-formedness in large language models: A comparative study with human judgments}.
\newblock In \emph{Proceedings of the Workshop on Cognitive Modeling and Computational Linguistics}, pages 189--198, Bangkok, Thailand. Association for Computational Linguistics.

\bibitem[{Qiu et~al.(2023)Qiu, Duan, and Cai}]{qiu_pragmatic_2023}
Zhuang Qiu, Xufeng Duan, and Zhenguang~Garry Cai. 2023.
\newblock \href {https://doi.org/10.31234/osf.io/qtbh9} {Pragmatic {Implicature} {Processing} in {ChatGPT}}.

\bibitem[{Qualtrics(2024)}]{qualtrics2024}
Qualtrics. 2024.
\newblock \href {https://www.qualtrics.com} {Qualtrics and all other qualtrics product or service names are registered trademarks or trademarks of qualtrics}.
\newblock Provo, UT, USA.

\bibitem[{Song et~al.(2024)Song, He, Jiang, Xian, Gao, Liu, and Yu}]{song-etal-2024-large}
Ran Song, Shizhu He, Shuting Jiang, Yantuan Xian, Shengxiang Gao, Kang Liu, and Zhengtao Yu. 2024.
\newblock \href {https://doi.org/10.18653/v1/2024.emnlp-main.403} {Does large language model contain task-specific neurons?}
\newblock In \emph{Proceedings of the 2024 Conference on Empirical Methods in Natural Language Processing}, pages 7101--7113, Miami, Florida, USA. Association for Computational Linguistics.

\bibitem[{Stella et~al.(2023)Stella, Hills, and Kenett}]{stella2023using}
Massimo Stella, Thomas~T Hills, and Yoed~N Kenett. 2023.
\newblock Using cognitive psychology to understand gpt-like models needs to extend beyond human biases.
\newblock \emph{Proceedings of the National Academy of Sciences}, 120(43):e2312911120.

\bibitem[{Tang et~al.(2024)Tang, Luo, Huang, Zhang, Wang, Zhao, Wei, and Wen}]{tang2024language}
Tianyi Tang, Wenyang Luo, Haoyang Huang, Dongdong Zhang, Xiaolei Wang, Xin Zhao, Furu Wei, and Ji-Rong Wen. 2024.
\newblock Language-specific neurons: The key to multilingual capabilities in large language models.
\newblock \emph{arXiv preprint arXiv:2402.16438}.

\bibitem[{Templeton(2024)}]{templeton2024scaling}
Adly Templeton. 2024.
\newblock \emph{Scaling monosemanticity: Extracting interpretable features from claude 3 sonnet}.
\newblock Anthropic.

\bibitem[{Vig et~al.(2020)Vig, Gehrmann, Belinkov, Qian, Nevo, Singer, and Shieber}]{vig2020investigating}
Jesse Vig, Sebastian Gehrmann, Yonatan Belinkov, Sharon Qian, Daniel Nevo, Yaron Singer, and Stuart Shieber. 2020.
\newblock Investigating gender bias in language models using causal mediation analysis.
\newblock \emph{Advances in neural information processing systems}, 33:12388--12401.

\bibitem[{Wang et~al.(2024{\natexlab{a}})Wang, Duan, and Cai}]{wang-etal-2024-multimodal}
Shuqi Wang, Xufeng Duan, and Zhenguang Cai. 2024{\natexlab{a}}.
\newblock \href {https://doi.org/10.18653/v1/2024.conll-1.32} {A multimodal large language model {``}foresees{''} objects based on verb information but not gender}.
\newblock In \emph{Proceedings of the 28th Conference on Computational Natural Language Learning}, pages 435--441, Miami, FL, USA. Association for Computational Linguistics.

\bibitem[{Wang et~al.(2024{\natexlab{b}})Wang, Haddow, Wu, Peng, and Birch}]{wang2024sharing}
Weixuan Wang, Barry Haddow, Minghao Wu, Wei Peng, and Alexandra Birch. 2024{\natexlab{b}}.
\newblock Sharing matters: Analysing neurons across languages and tasks in llms.
\newblock \emph{arXiv preprint arXiv:2406.09265}.

\bibitem[{Warstadt et~al.(2019)Warstadt, Singh, and Bowman}]{warstadt_neural_2019}
Alex Warstadt, Amanpreet Singh, and Samuel~R. Bowman. 2019.
\newblock \href {https://doi.org/10.1162/tacl_a_00290} {Neural {Network} {Acceptability} {Judgments}}.
\newblock \emph{Transactions of the Association for Computational Linguistics}, 7:625--641.
\newblock Place: Cambridge, MA Publisher: MIT Press.

\bibitem[{Wei et~al.(2022)Wei, Tay, Bommasani, Raffel, Zoph, Borgeaud, Yogatama, Bosma, Zhou, Metzler, Chi, Hashimoto, Vinyals, Liang, Dean, and Fedus}]{wei_emergent_2022}
Jason Wei, Yi~Tay, Rishi Bommasani, Colin Raffel, Barret Zoph, Sebastian Borgeaud, Dani Yogatama, Maarten Bosma, Denny Zhou, Donald Metzler, Ed~H. Chi, Tatsunori Hashimoto, Oriol Vinyals, Percy Liang, Jeff Dean, and William Fedus. 2022.
\newblock \href {https://doi.org/10.48550/arXiv.2206.07682} {Emergent {Abilities} of {Large} {Language} {Models}}.
\newblock \emph{arXiv preprint}.
\newblock ArXiv:2206.07682 [cs].

\bibitem[{Zhang et~al.(2024)Zhang, Zhao, Zhang, Gui, and Huang}]{zhang_unveiling_2024}
Zhihao Zhang, Jun Zhao, Qi~Zhang, Tao Gui, and Xuanjing Huang. 2024.
\newblock \href {https://doi.org/10.48550/arXiv.2402.14700} {Unveiling {Linguistic} {Regions} in {Large} {Language} {Models}}.
\newblock \emph{arXiv preprint}.
\newblock ArXiv:2402.14700 [cs].

\bibitem[{Zhou et~al.(2024)Zhou, Chen, Cahyawijaya, Duan, and Cai}]{zhou2024linguistic}
Xinyu Zhou, Delong Chen, Samuel Cahyawijaya, Xufeng Duan, and Zhenguang~G Cai. 2024.
\newblock Linguistic minimal pairs elicit linguistic similarity in large language models.
\newblock \emph{arXiv preprint arXiv:2409.12435}.

\end{thebibliography}

\end{document}